\documentclass[letterpaper, 10 pt, conference]{ieeeconf}  %

\IEEEoverridecommandlockouts                              %

\usepackage{amsmath} %
\usepackage{amssymb}  %
\usepackage{color}
\usepackage{siunitx}
\usepackage{hhline}
\usepackage{tikz}
\usepackage{graphicx} %

\let\labelindent\relax
\usepackage{enumitem}

\usepackage{booktabs}
\usepackage{multirow}
\usepackage{xifthen}

\usepackage[noadjust]{cite}

\makeatletter
\let\NAT@parse\undefined
\makeatother
\usepackage[bookmarks=false]{hyperref}

\usepackage{fancyhdr}
\usetikzlibrary{positioning}
\newcommand\updateRate[3]{
    \begin{tikzpicture}[every node/.style={inner sep=0,outer sep=0}]]
        \node[anchor=east] at (0, 0) (a1) {#1};%
        \node[above right=-0.01 and 0.45 of a1, anchor=east, align=right, font=\tiny, color=gray] (a2) {#2};%
        \node[below right=0.1 and 0.0 of a2, anchor=east, align=right, font=\tiny, color=gray] (a3) {#3};%
    \end{tikzpicture}
}

\title{\LARGE \bf
Efficient and Robust Semantic Mapping for Indoor Environments%
}

\author{Daniel Seichter, Patrick Langer, Tim Wengefeld, Benjamin Lewandowski, Dominik Höchemer, and \\Horst-Michael Gross%
\thanks{Authors are with Neuroinformatics and Cognitive Robotics Lab, Technische Universit\"at Ilmenau, 98693 Ilmenau, Germany. Contact: \hfill\newline
{\scriptsize{\tt daniel.seichter@tu-ilmenau.de}, ORCID: {\tt
\href{https://orcid.org/0000-0002-3828-2926}{0000-0002-3828-2926}}}}%
\thanks{This work has received funding from the German Federal Ministry of Education and Research (BMBF) to the project MORPHIA (16SV8426).}%
}

\begin{document}

\maketitle

\newboolean{isarxiv}
\setboolean{isarxiv}{true}
\ifthenelse{\boolean{isarxiv}}{%
    \renewcommand{\headrulewidth}{0pt}
    \fancypagestyle{fancyfirstpage}{%
        \fancyhf{}%
        \fancyhead[C]{%
            \footnotesize%
            \vspace{-15mm}%
            \textcolor{gray}{%
                © 2022 IEEE.  Personal use of this material is permitted.  Permission from IEEE must be obtained for all other uses, in any current or future media, including reprinting/republishing this material for advertising or promotional purposes, creating new collective works, for resale or redistribution to servers or lists, or reuse of any copyrighted component of this work in other works.
            }%
        }
        \fancyfoot[C]{%
            \footnotesize%
            \textcolor{gray}{\thepage}%
        }
    }
    \fancypagestyle{fancypage}{%
        \fancyhf{}%
        \fancyfoot[C]{%
            \footnotesize%
            \textcolor{gray}{\thepage}%
        }        
    }    
    \thispagestyle{fancyfirstpage}
    \pagestyle{fancypage}
}{%
    \thispagestyle{empty}%
    \pagestyle{empty}%
}%

\begin{abstract}
A key proficiency an autonomous mobile robot must have to perform high-level tasks is a strong understanding of its environment.
This involves information about what types of objects are present, where they are, what their spatial extend is, and how they can be reached, i.e., information about free space is also crucial. 
Semantic maps are a powerful instrument providing such information.
However, applying semantic segmentation and building 3D maps with high spatial resolution is challenging given limited resources on mobile robots.
In this paper, we incorporate semantic information into efficient occupancy normal distribution transform (NDT) maps to enable real-time semantic mapping on mobile robots.
On the publicly available dataset Hypersim, we show that, due to their sub-voxel accuracy, semantic NDT maps are superior to other approaches.
We compare them to the recent state-of-the-art approach based on voxels and semantic Bayesian spatial kernel inference~(S-BKI) and to an optimized version of it derived in this paper.
The proposed semantic NDT maps can represent semantics to the same level of detail, while mapping is 2.7 to 17.5 times faster.
For the same grid resolution, they perform significantly better, while mapping is up to more than 5 times faster.
Finally, we prove the real-world applicability of semantic NDT maps with qualitative results in a domestic application.
\end{abstract}%

\section{Introduction}
\label{sec:introduction}
As the application scenarios of mobile robots are getting more complex and challenging, semantic scene understanding becomes increasingly crucial.
A robot always needs to understand its environment including the significance of a scene and all objects within. 
One way to accomplish this scene understanding is to build a semantic 3D representation of the environment through mapping~\cite{Sengupta-ICRA-2015, Yang-IROS-2017, Jadidi-ArXiV-2017, Gan-RAL-2020} and pass it to subsequent analysis modules.

Building on previous own work~\cite{Gross-IROS-2015,Gross-ICRA-2019}, in our ongoing research project MORPHIA, our robots are acting autonomously in domestic environments and allow relatives as well as caregivers to stay in contact with people in need of care in order to let them participate in social life.
In this scenario, scene understanding gained from a semantic 3D representation is supposed to improve navigation and to enable semantic instructions such as ``wait for me next to the chair''.
Moreover, it is to allow our robots to be aware of the current scene, i.e., by not blocking doors or the line of sight to objects such as a TV.
To match the character of these dynamic environments, we build upon a strong long-term localization determined via RTAB-Map~\cite{Labbe-JFR-2019} and build short-term semantic 3D representations of the current environment periodically.
This way, we are able to react to small but important changes in the environment, such as rearranged chairs, clothes, or other dynamic objects.

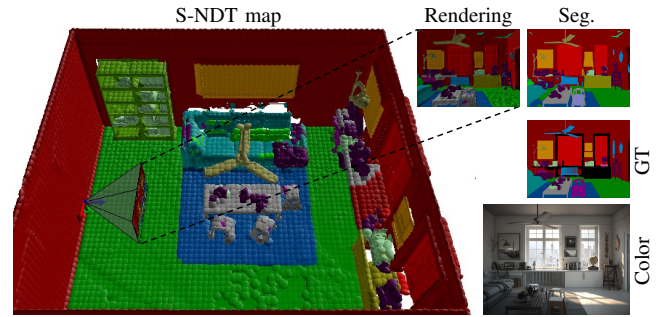
\begin{figure}[!t]
    \vspace{0.5mm}
	\centering
	\begin{tikzpicture}[scale=1.0]
	    \node at (0, 0){%
	        \includegraphics[width=0.95\columnwidth, trim=1.16cm 8.5cm 10.2cm 0cm, clip]{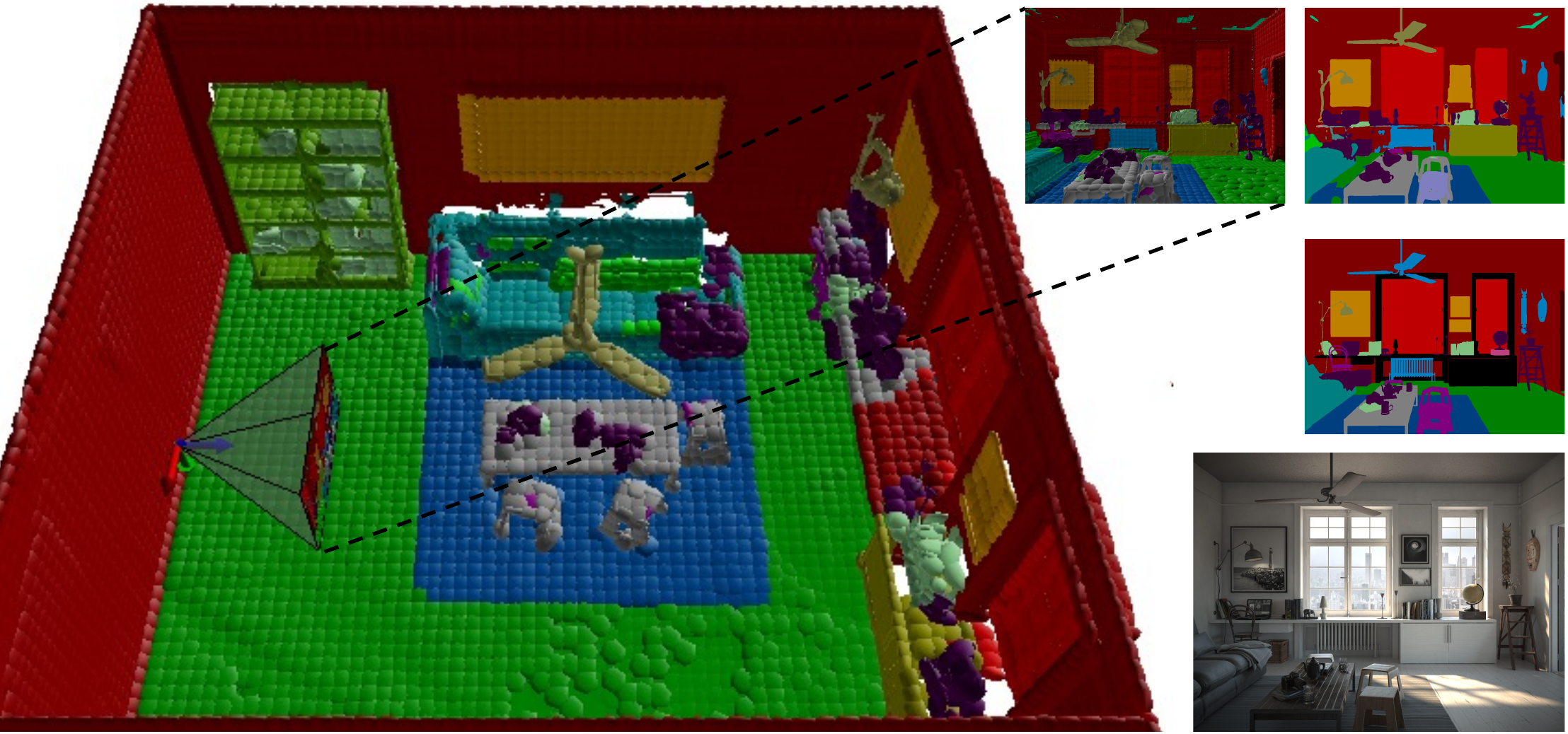}
	    };
	    \node[rotate=90] at (4.25, -1.2){\footnotesize Color};%
	    \node[rotate=90] at (4.25, 0.13){\footnotesize GT};%
	    \node at (-1.25, 2.05){\footnotesize S-NDT map};%
	    \node[anchor=east] at (2.65, 2.05){\footnotesize Rendering};%
	    \node[anchor=east] at (3.75, 2.05){\footnotesize Seg.};%
	\end{tikzpicture}%
	\vspace{-4mm}
    \caption{%
        Semantic occupancy NDT (S-NDT) map built with predicted segmentation of ESANet~\cite{Seichter-ICRA-2021-ESANet} and cell size of 10\si{\centi\meter} for scene \emph{ai\_048\_010} of the Hypersim~\cite{Roberts-ICCV-2021-Hypersim} test set. %
        Right: for the given pose counterclockwise from bottom: color image, ground-truth annotation (GT), semantic segmentation output of ESANet (Seg.), and 2D rendering visualizing the map from the robot's pose (Rendering). %
        Best viewed in color at 300\%. %
        Black indicates the void class, see Fig.~\ref{fig:experiments:ious} for remaining colors.%
    }%
    \label{fig:eyecatcher}
    \vspace{-4mm}
\end{figure}

In the field of mobile robotics, voxel-based 3D representations are most common due to their efficient processing~\cite{Hornung-AR-2013, Wang-ICRA-2016, Doherty-ICRA-2017}.
Approaches for creating semantic maps often extend these classical mapping approaches by additionally incorporating semantic information~\cite{Sengupta-ICRA-2015, Yang-IROS-2017, Jadidi-ArXiV-2017, Gan-RAL-2020}.
However, none of the aforementioned approaches is able to meet our requirements for efficient and robust semantic mapping.

Therefore, in this paper, we aim at developing an approach that can be used in our scenario.
We first propose to modify the update step of the recently published semantic Bayesian kernel inference mapping approach~(S-BKI)~\cite{Gan-RAL-2020} in order to improve its robustness when it is combined with noisy outputs of a semantic segmentation.
Unfortunately, this optimized version still does not allow precise semantic mapping at a frame rate of at least 3\si{\hertz} on our mobile robot.
Therefore, we also propose to integrate semantic information into the fast and robust occupancy normal distribution transform (NDT) mapping of~\cite{Einhorn-ECMR-2013-NDT, Einhorn-PHD-2019-NDT}, enabling precise semantic mapping at sub-voxel level as shown in Fig.~\ref{fig:eyecatcher}.

Evaluating semantic mapping approaches for indoor environments is challenging as the number of suitable datasets is limited.
In order to compare both proposed approaches to the state of the art, we conduct experiments on the synthetic large-scale dataset Hypersim~\cite{Roberts-ICCV-2021-Hypersim}.
We first evaluate their capability to represent semantics by using the provided ground-truth annotation for mapping, i.e., perfect data is assumed.
Subsequently, we combine the mapping with our efficient RGB-D semantic segmentation approach~\cite{Seichter-ICRA-2021-ESANet} in order to allow a comparison in a more realistic setting when semantic information is noisy.
Finally, we show qualitative results of our complete pipeline for efficient and robust semantic mapping in one of our domestic environments and, thus, give an impression of how it performs on real data.
In summary, the main contributions of this paper are:
\begin{itemize}[leftmargin=4mm]
    \item a modified update step for the semantic Bayesian kernel inference (S-BKI) mapping~\cite{Gan-RAL-2020} improving its robustness when it is combined with noisy outputs of a segmentation
    \item the integration of semantic information into the occupancy normal distribution transform (NDT) mapping~\cite{Einhorn-ECMR-2013-NDT, Einhorn-PHD-2019-NDT} realizing semantic NDT maps
    \item a comparison of the aforementioned approaches on the synthetic large-scale dataset Hypersim~\cite{Roberts-ICCV-2021-Hypersim}.
\end{itemize}
\section{Related Work}
\label{sec:related_work}
Representations for 3D environments are diverse, ranging from raw point clouds over meshes~\cite{Zhou-MMM-2017} and voxels~\cite{Hornung-AR-2013} to implicit descriptor-based~\cite{Shin-CVPR-2018} representations. 
However, in the field of mobile robotics, representations are most commonly based on voxels due to their efficient processing by subsequent algorithms and controllable need for memory.

\subsection{Voxel-based Data Representations}
\label{sec:related_work:mapping}
Voxel-based approaches divide the space into a regular 3D grid of cells with a specific size.
These cells can be organized in an octree.
OctoMap~\cite{Hornung-AR-2013}, for example, builds a 3D map by estimating the status of a voxel either as occupied, free, or unknown given range measurements of a sensor.
However, when point density is low, e.g., due to large distances, 
na\"ive map updates with raw sensor readings may lead to holes in the final map.
This issue is targeted by Gaussian processes in GPOctoMap~\cite{Wang-ICRA-2016} or kernel-based approaches such as Bayesian generalized kernel OctoMap~(BGKOctoMap)~\cite{Doherty-ICRA-2017}. 
Nevertheless, the map precision of these approaches strongly depends on the resolution of the 3D grid.
As the cell size also affects the required memory and computation time, a trade-off must be found.
Therefore, other approaches do not only estimate an occupancy value per voxel but also attempt to model the surface within the cells.
This allows using coarser grid resolutions and, thus, retains real-time requirements while still modelling accurate representations of the environment.
Surfel maps~\cite{Klass-GCR-2012}, for example, use so called surface elements that are characterized by the eigenvectors of the data distributed inside a voxel.
However, this representation makes use of a flat surface assumption that becomes insufficient when modelling objects smaller than the grid resolution.
A similar representation is the normal distribution transform (NDT)~\cite{Biber-IROS-2003-NDT,Magnusson-JFR-2007-NDT,Einhorn-ECMR-2013-NDT} that maintains the whole covariance matrix of the distribution, enabling to model the spatial extend inside a voxel.

\subsection{Semantic Voxel-based Data Representations}
\label{sec:related_work:semantic_mapping}
When semantic information has to be integrated into grid-based environment models, approaches typically extend non-semantic approaches.
Semantic Octree~\cite{Sengupta-ICRA-2015} extends OctoMap by integrating semantic information using a class histogram per voxel.
Semantic labels are retrieved from RGB images segmented using a conditional random field (CRF).
Finally, the map is further refined using a higher-order CRF.
In~\cite{Yang-IROS-2017}, this approach is improved by incorporating a convolutional neural network for semantic segmentation.
\cite{Jadidi-ArXiV-2017} extends the idea of using Gaussian processes for inferring not only a binary label (free$\,$vs.$\,$occupied) but a label in a multi-class problem (free$\,$+$\,$semantic classes).
However, they do not make use of an octree-based data structure like GPOctoMap, which makes this approach less adaptable for larger environments.
Semantic Bayesian kernel inference mapping~(S-BKI)~\cite{Gan-RAL-2020} as generalization of BGKOctoMap models a multi-class problem similar to~\cite{Jadidi-ArXiV-2017} but using faster kernel-based inference.
However, the free space class, which is sampled from range measurements, and any other semantic information from the segmentation are handled in the same way.
This can lead to over-representing the free space class and, thus, to inconsistencies in the map, especially when combined with noisy outputs of a semantic segmentation.

Another related field of research is 3D scene reconstruction.
Approaches such as \cite{Mccormac-ICRA-2017,jeon2018semantic,pham2019real,narita-IROS-2019} reconstruct entire semantic scenes using volumetric mapping based on voxel hashing~\cite{niessner2013real} and truncated signed distance fields~(TSDF)~\cite{curless1996volumetric}. 
However, for efficiency reasons, TSDFs explicitly model free or occupied space only in a truncated region around the surface of objects. 
Information about unseen areas are missing.
Moreover, these approaches rely on computationally intensive CRFs for map refinement that require to be heavily skipped in application to enable at least near real-time capability even on non-mobile CPUs and GPUs.
Thus, these approaches are not suitable for mobile robotic applications with dynamic environments and limited resources.

In conclusion, integrating semantic information into 3D grid-based data structures has been targeted by several approaches.
However, to the best of our knowledge, none of these approaches make use of the advanced NDT data representation that is able to model sub-voxel precise semantic information accurately.
For the sake of completeness, in~\cite{Zaganidis-IROS-2017}, semantics has already been used to improve the scan registration process by building class-specific NDT maps but no consistent semantic global map is generated. 
\section{Efficient and Robust Semantic Mapping}
\label{sec:main}

\begin{figure}[!t]%
	\vspace{2.5mm}
    \centering
        \includegraphics[width=0.95\columnwidth, trim=0 2.9cm 1.2cm 0, clip]{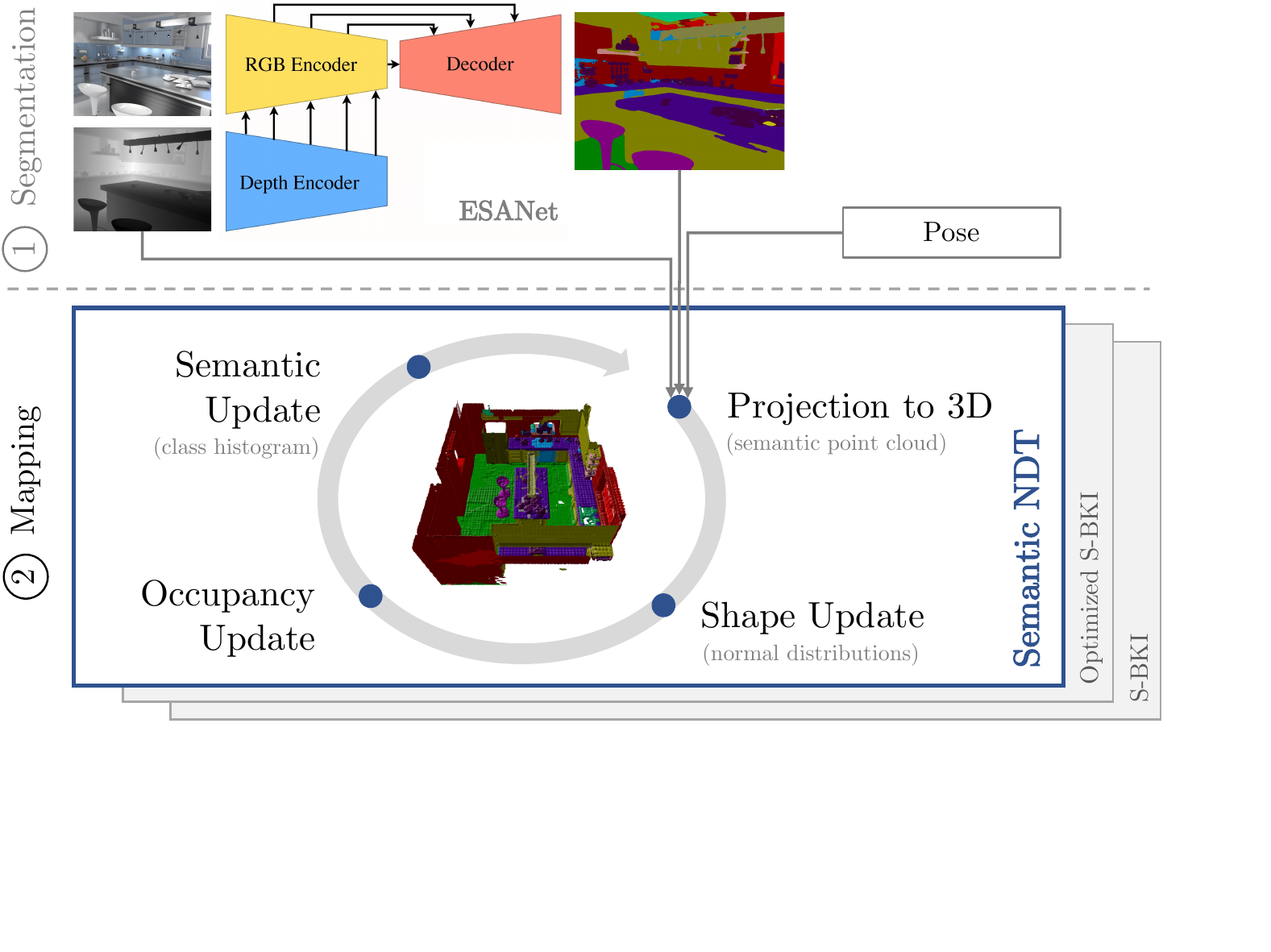}%
    \vspace{-3mm}
	\caption{%
	    Our two-step approach for efficient and robust semantic mapping.
	}
	\vspace{-3mm}
	\label{fig:approach}
\end{figure}

Given a precise localization in the environment, our semantic mapping pipeline is a two-step approach as shown in Fig.~\ref{fig:approach}. 
We first apply a semantic segmentation on the current set of images~(color and depth).
Afterwards, the obtained per-pixel class information are passed together with the depth image and the current pose to the mapping stage.
Unlike other approaches~\cite{Yang-IROS-2017,Mccormac-ICRA-2017,jeon2018semantic,pham2019real,narita-IROS-2019}, we do not apply any post-processing to the resulting maps, such as CRFs, as this would notably increase runtime.
For semantic segmentation, we rely on our recently proposed RGB-D approach for efficient indoor scene analysis, called ESANet~\cite{Seichter-ICRA-2021-ESANet}.
Semantic mapping can be performed using either semantic Bayesian kernel inference mapping (S-BKI), the proposed optimized version of this approach~(OS-BKI), or the proposed semantic occupancy normal distribution transform mapping (S-NDT).
Next, we describe each part of the pipeline in detail.

\subsection{Semantic Segmentation}
\label{sec:main:segmentation}
Unlike most other approaches~\cite{Sengupta-ICRA-2015,Yang-IROS-2017, Jadidi-ArXiV-2017, Gan-RAL-2020, pham2019real, narita-IROS-2019}, we follow~\cite{Mccormac-ICRA-2017, jeon2018semantic} and use an RGB-D approach for semantic segmentation.
Especially for indoor environments, cluttered scenes may impede  semantic segmentation.
Incorporating depth images can alleviate this effect by providing complementary geometric information.
In~\cite{Seichter-ICRA-2021-ESANet}, we have shown that RGB-D segmentation is superior to processing RGB images solely and that it can still be performed in real time if the network architecture is carefully designed.
More precisely, we use ESANet-R34-NBt1D (enhanced ResNet34-based encoder utilizing the Non-Bottleneck-1D block~(NBt1D)~\cite{ERFNet-its2018}) in our pipeline.
It enables semantic segmentation with up to $29.7\si{\hertz}$ on a NVIDIA Jetson AGX Xavier making it well suited for efficient and robust semantic mapping on mobile robots.
For application in real-world scenarios, pretrained weights for both datatsets NYUv2~\cite{Silverman-ECCV-2012-NYUv2} and SUNRGB-D~\cite{Song-CVPR-2015-SUNRGBD} are publicly available, reaching a mean intersection over union of 51.58 and 48.17 on the test sets, respectively. 

\subsection{Mapping: Optimized S-BKI (OS-BKI)}
\label{sec:main:opt_sbki}

S-BKI~\cite{Gan-RAL-2020} extends BGKOctoMap~\cite{Doherty-ICRA-2017} by additionally incorporating semantic labels.
Both use Bayesian generalized kernel inference\footnote{For further precise mathematical descriptions, we refer to~\cite{Gan-RAL-2020, Doherty-ICRA-2017}.} for determining the label of a voxel in order to improve mapping in regions with low point density.
The label is inferred by weighting all observations up to a specified maximum distance to the current voxel~(kernel length~$l$) using a kernel function.
Since this does not consider information about free space inherently, additional points labeled as free are sampled along the sensor's rays and are incorporated into the kernel inference.

\begin{figure}[!b]%
    \vspace{-3mm}
	\centering%
	\begin{tikzpicture}[scale=1.0]%
	    \node[anchor=north west] at (0, 0){%
	        \includegraphics[width=3.9cm,trim=1cm 5cm 3cm 3cm, clip]{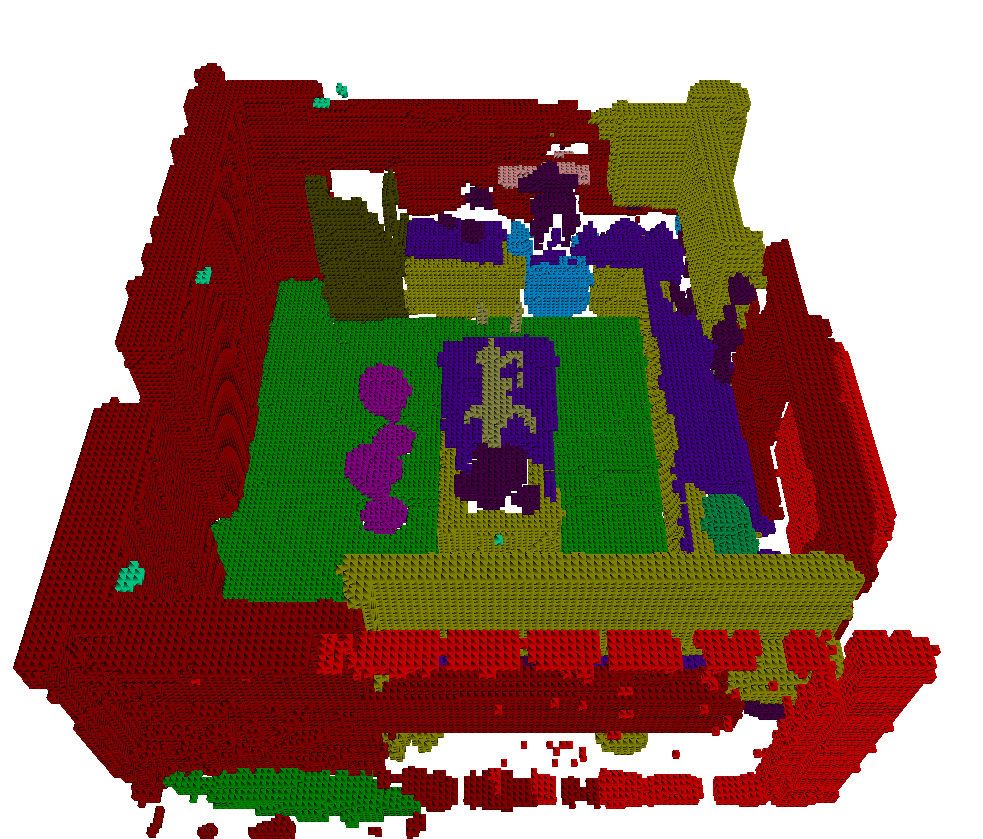}%
	    };
	    \node[anchor=north west] at (4.25, 0){%
	        \includegraphics[width=3.9cm, trim=1cm 5cm 3cm 3cm, clip]{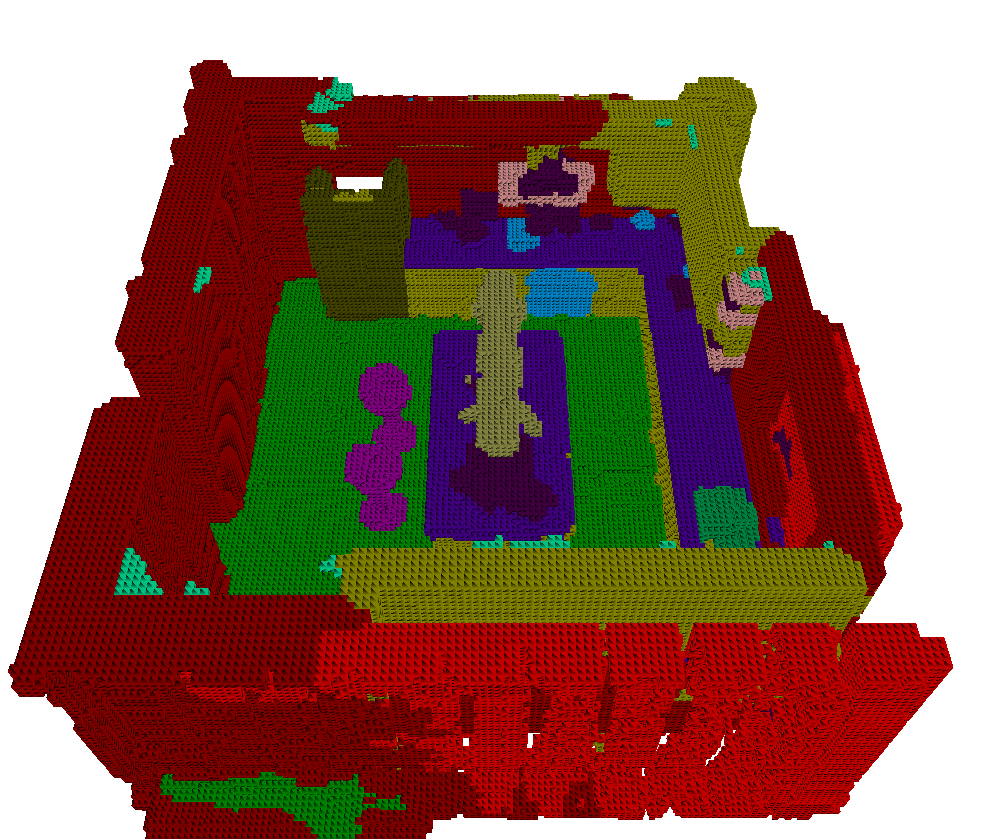}%
	    };
	    \node[anchor=center] at (2.075,-3.15) {\footnotesize S-BKI~\cite{Gan-RAL-2020}};
	    \node[anchor=center] at (6.35,-3.15) {\footnotesize OS-BKI (ours)};
	\end{tikzpicture}%
	\vspace{-4mm}
	\caption{%
	    Semantic maps built with S-BKI~(left) and OS-BKI~(right) for scene \emph{ai\_001\_010} of the Hypersim test set. 
	    Mapping was done with predicted segmentation of ESANet~\cite{Seichter-ICRA-2021-ESANet}, cell size of 5\si{\centi\meter}, and kernel length of 20\si{\centi\meter}. %
	    Best viewed in color at 200\%, see Fig.~\ref{fig:experiments:ious} for label colors.}
	\label{fig:bki_problem}
	\vspace{-1mm}
\end{figure}

However, S-BKI distinguishes 1$\,$+$\,$n classes (free$\,$+$\,$n semantic classes) in the same way as BGKOctoMap only distinguishes free and occupied.
As shown in Fig.~\ref{fig:bki_problem}~(left), this leads to major problems in terms of holes in the map.
Especially when combining with noisy outputs of a semantic segmentation, the frequency for the free class tends to be the highest in areas where multiple semantic labels occur.
This is due to the fact that each semantic class is counted separately, although all semantic observations indicate occupancy.
A larger kernel length or an even more noisy segmentation further increase this effect.

To tackle this issue, we propose to split the update in two consecutive steps.
In the first step, the kernel is applied to a two-class problem ignoring the semantic at all and distinguishing only free and occupied as done in BGKOctoMap.
The second step is used to infer semantic labels but only considering semantic observations.
Although applying the kernel two times sounds like more overhead, the computation time is only slightly affected as shown in our experiments.
This is due to the fact that the complexity of the first step and the number of observations in the second step are reduced.
As shown in Fig.~\ref{fig:bki_problem}~(right) and our experimental results, OS-BKI greatly improves mapping quality.

\subsection{Mapping: Semantic Occupancy NDT (S-NDT)}
\label{sec:main:sdnt}
The proposed OS-BKI approach improves mapping with noisy outputs of a semantic segmentation.
Nevertheless, mapping with sufficient resolution is not possible in real time on our mobile robot without relying on multiple CPU cores.
However, saving computational resources is important since our mobile robot typically has to perform other tasks in parallel, like navigation or person perception and tracking.

Therefore, we propose semantic occupancy NDT~(S-NDT) mapping that allows efficient semantic mapping and, moreover, representing the environment on a sub-voxel level.
Compared to traditional voxel maps~\cite{Hornung-AR-2013}, 3D occupancy NDT maps not only store an occupancy value per voxel but also a three-dimensional mean vector and a $3{\times}3$ covariance matrix, describing the underlying surface in terms of a normal distribution.
These normal distributions can be updated efficiently in an incremental fashion as shown in~\cite{Einhorn-ECMR-2013-NDT}.
We integrate semantic information by additionally introducing a class histogram in each voxel. 
This histogram is updated using the top-k labels of the segmentation output.
However, in accordance with other approaches~\cite{Jadidi-ArXiV-2017,Gan-RAL-2020}, the top-1 labels already seem to be sufficient.

\begin{figure}[!b]
	\centering
	\begin{tikzpicture}[scale=1.0]%
	    \node[anchor=north west] at (0, 0){%
	        \includegraphics[width=3.9cm,trim=1cm 5cm 3cm 3cm, clip]{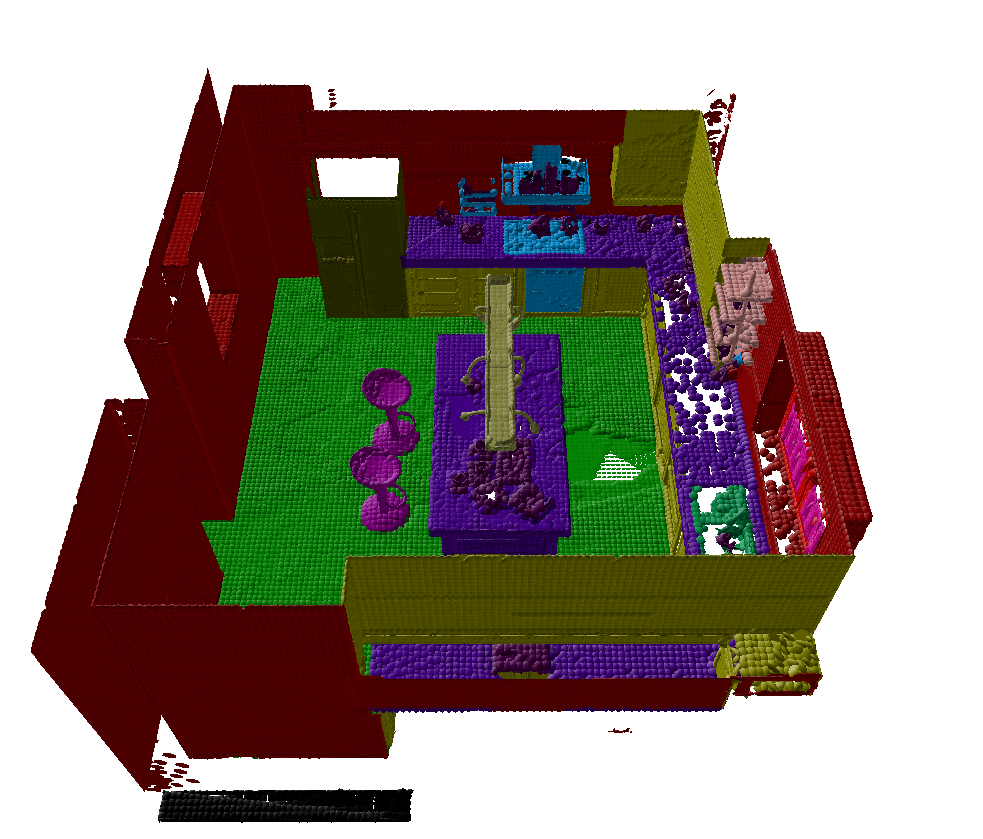}%
	    };
	    \node[anchor=north west] at (4.15, 0){%
	        \includegraphics[width=3.9cm, trim=1cm 5cm 3cm 3cm, clip]{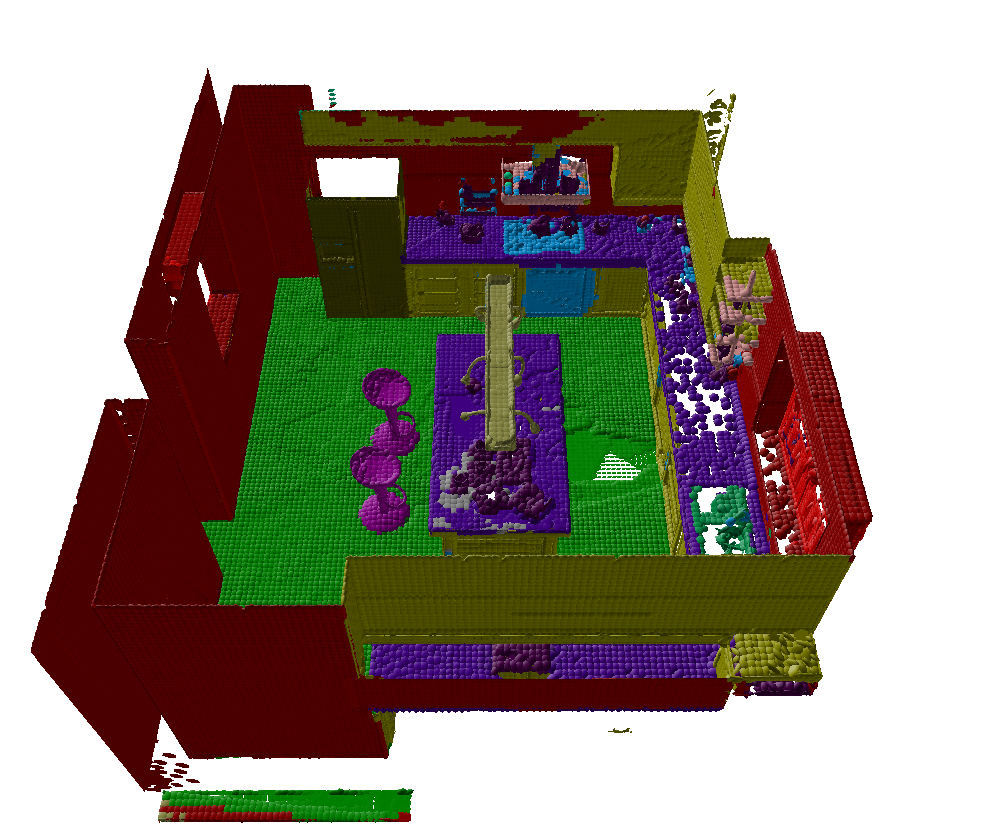}%
	    };
	    \node[anchor=north west] at (0, -3.0){%
	        \includegraphics[width=3.9cm, trim=1cm 5cm 3cm 3cm, clip]{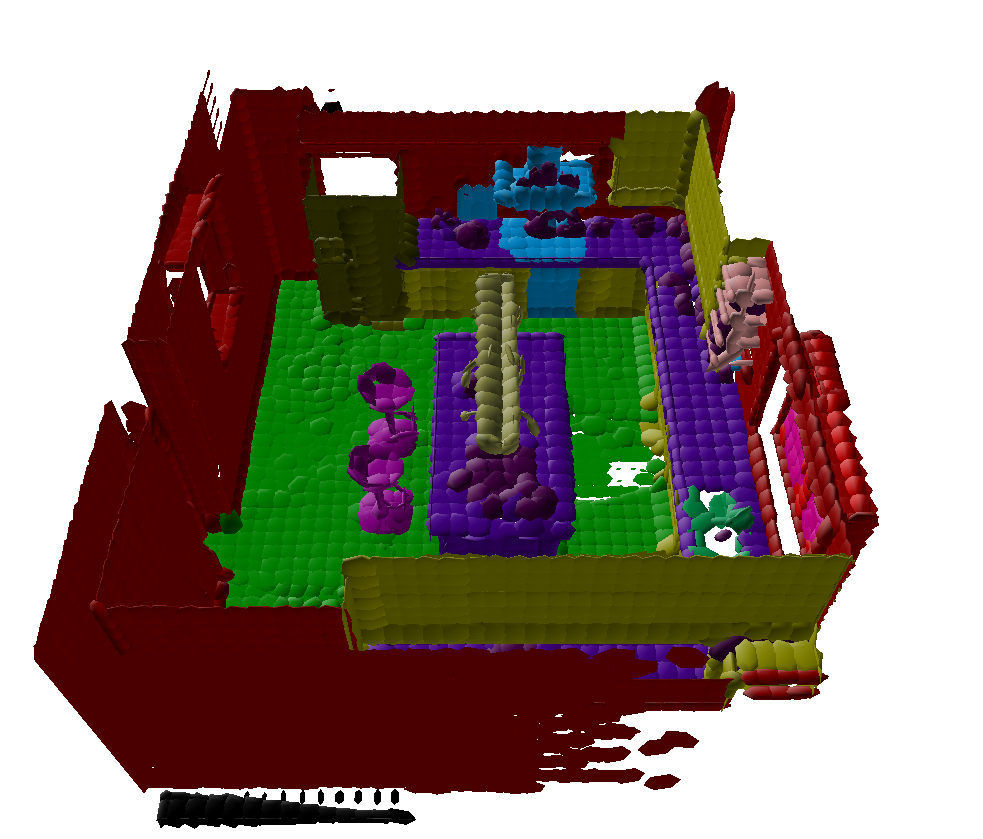}%
	    };
	    \node[anchor=north west] at (4.15, -3.0){%
	        \includegraphics[width=3.9cm, trim=1cm 5cm 3cm 3cm, clip]{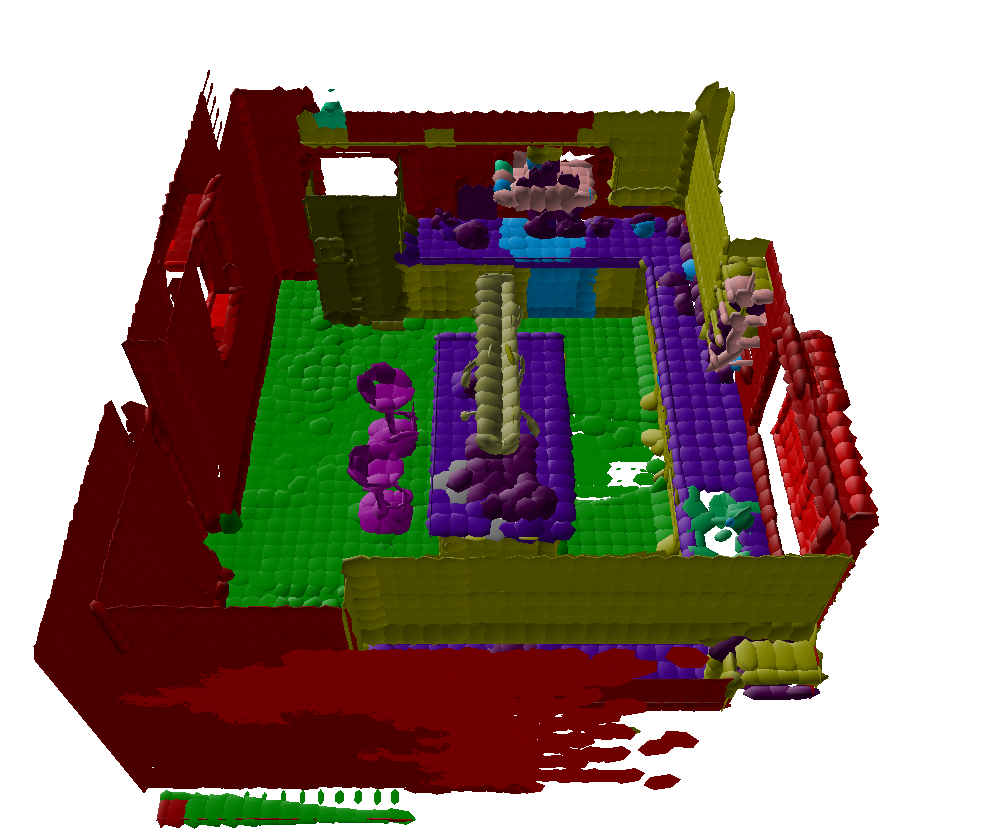}%
	    };	    
	    \node[rotate=90, anchor=center] at (-0.125,-1.45) {\footnotesize S-NDT (5\si{\centi\meter})};
	    \node[rotate=90, anchor=center] at (-0.125,-4.55) {\footnotesize S-NDT (15\si{\centi\meter})};
	    \node[anchor=center] at (2.075,-6.125) {\footnotesize Ground-truth segmentation};
	    \node[anchor=center] at (6.25,-6.125) {\footnotesize Predicted segmentation};
	\end{tikzpicture}%
	\vspace{-4mm}
    \caption{%
        Semantic maps built with S-NDT with cell sizes of 5\si{\centi\meter}~(top) and 15\si{\centi\meter} (bottom) for scene \emph{ai\_001\_010} of the Hypersim test set. %
        Columns show results when mapping with ground-truth segmentation~(left, perfect input for mapping) or predicted segmentation of ESANet~\cite{Seichter-ICRA-2021-ESANet}~(right, noisy input for mapping). %
        Best viewed in color at 200\%, see Fig.~\ref{fig:experiments:ious} for label colors.}%
    \label{fig:ndt}%
    \vspace{-1mm}
\end{figure}
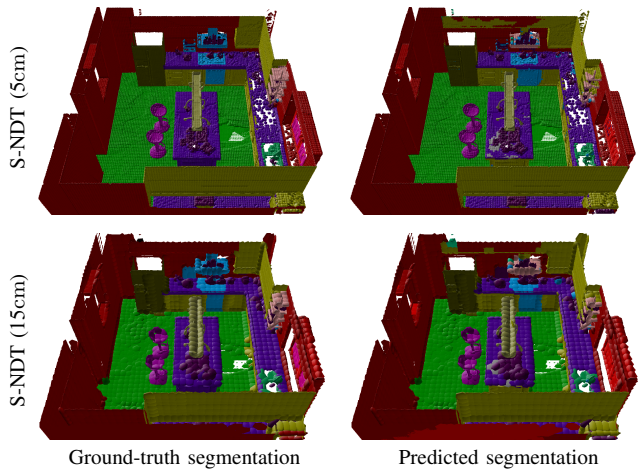

In \cite{Stoyanov-ICRA-2011-NDT,Einhorn-ECMR-2013-NDT, Einhorn-PHD-2019-NDT}, it is shown that occupancy NDT maps have a higher effective resolution, i.e., they are able to represent a surface more accurately at the same grid resolution due to their sub-voxel surface description.
In other words, occupancy NDT maps can achieve a similar accuracy with a reduced grid resolution.
This does not only save computational resources during mapping but also enlarges the spatial volume that is covered per voxel.
As shown with the countertop on the right in Fig.~\ref{fig:ndt}, this also compensates for input data with low point density, since the resulting map is less incomplete.
In large parts, the semantic occupancy NDT map with cell size 15\si{\centi\meter} (Fig.~\ref{fig:ndt}, bottom) is similarly accurate as its OS-BKI counterpart with 5\si{\centi\meter}~(Fig.~\ref{fig:bki_problem}, right).

\section{Experiments}
\label{sec:experiments}
Evaluating semantic mapping approaches is challenging.
Especially for indoor environments, the number of datasets is limited.
NYUv2~\cite{Silverman-ECCV-2012-NYUv2} and SUNRGB-D~\cite{Song-CVPR-2015-SUNRGBD} are used for benchmarking semantic segmentation approaches.
However, they only provide semantics for single frames of a scene and, thus, are not suitable for evaluating semantic mapping.
Therefore, we conduct mapping experiments on the recently published synthetic large-scale Hypersim dataset~\cite{Roberts-ICCV-2021-Hypersim}.

\subsection{Dataset \& Network Training}
\label{sec:experiments:data_network}
The Hypersim dataset~\cite{Roberts-ICCV-2021-Hypersim} contains 77,400~images of 461~photo-realistic synthetic indoor scenes (see Fig.~\ref{fig:eyecatcher} bottom right).
In total, 774~camera trajectories have been rendered using random walks for these scenes.
Each trajectory consists of 100 camera poses, ensuring reasonable visual coverage.
For each camera pose, among others, camera extrinsics, color and depth image as well as per-pixel semantic labels following the NYUv2 classes are available.
The dataset comes with splits for training, validation, and test.
We used the training set with a random subsampling of~20\% per epoch for training our ESANet-R34-NBt1D~\cite{Seichter-ICRA-2021-ESANet}.
The validation set was used to determine the best epoch.
All mapping experiments are conducted on the test set.
Unfortunately, the dataset contains some invalid scenes and trajectories~(void/single label only, missing textures, invalid depth) that we filtered from training and validation set manually.
However, although containing invalid scenes as well, we did not touch the test set.

\subsection{Implementation Details \& Evaluation Protocol}
\label{sec:experiments:implementation_evaluation}
Since each trajectory ensures reasonable visual coverage of the respective scene, we map each trajectory independently in our experiments.
We use either the ground-truth annotation or the output of our ESANet-R34-NBt1D as segmentation labels. 
Similar to \cite{Yang-IROS-2017, Gan-RAL-2020}, we limit the maximum mapping distance in our experiments to a reasonable value of 20\si{\meter}.
For all approaches, we evaluate cell sizes of 5, 10, 15, and 20\si{\centi\meter}.
For S-BKI and OS-BKI, we vary the kernel length~(denoted in brackets) in steps of 10\si{\centi\meter} between 10\si{\centi\meter} and 40\si{\centi\meter}.

For evaluating the resulting semantic maps, we follow~\cite{Yang-IROS-2017, Gan-RAL-2020}. 
We project the semantic map back to the camera plane for all camera poses of a trajectory and compute the mean intersection over union (mIoU) with the ground-truth annotation ignoring pixels labeled as void in the ground truth and classes that do not appear in the test set at all. 
However, the back-projections can include another class of invalid pixels, i.e., pixels indicating free space and no semantic class~(see Fig.~\ref{fig:experiments:backprojection}).
As the IoUs can only account for these results as false negatives and not as false positives for this invalid class (there is no ground truth), the mIoU might be misleading.
Therefore, we further report the overall ratio of invalid pixels~(invR) and the mean pixcel accuracy~(mPAcc) considering only true positives and false negatives per class.

In order to enable the community to have a similar setup, our code for preparing the Hypersim dataset, the weights for our ESANet-R34-NBt1D reaching a mIoU of 41.17 on the Hypersim test set, and the code for evaluation are publicly available at:
\href{https://github.com/TUI-NICR/semantic-mapping}{\small \url{https://github.com/TUI-NICR/semantic-mapping}}.

\subsection{Mapping with Perfect Ground-truth Segmentation}
\label{sec:experiments:gt}

\begin{table*}[!t]
\caption{%
    Results on the Hypersim test set when mapping with ground-truth~(GT) segmentation~(top) and predicted segmentation of ESANet~\cite{Seichter-ICRA-2021-ESANet}~(bottom). %
    For each configuration, four cell sizes are considered, and the mean intersection over union (mIoU), the ratio of invalid pixels (invR), and the mean pixel accuracy~(mPAcc) over all trajectories are listed as percentages. %
    For S-BKI and OS-BKI, the kernel length is denoted in brackets. %
    For each cell size, the best result for mIoU and mPAcc is printed in bold.
}
\vspace{-2.5mm}
\centering
\resizebox{0.96\textwidth}{!}{%

\begin{tabular}{@{ }l@{\hspace{2mm}}l@{\hspace{7mm}}c@{\hspace{1.5mm}}c@{\hspace{1mm}}c@{\hspace{7mm}}c@{\hspace{1.5mm}}c@{\hspace{1mm}}c@{\hspace{7mm}}c@{\hspace{1.5mm}}c@{\hspace{1mm}}c@{\hspace{7mm}}c@{\hspace{1.5mm}}c@{\hspace{1mm}}c@{}}
\toprule[1pt]%
                  &               & \multicolumn{3}{c@{\hspace{10mm}}}{\textbf{Cell size 5cm}}                                   & \multicolumn{3}{c@{\hspace{9mm}}}{\textbf{Cell size 10cm}}                                  & \multicolumn{3}{c@{\hspace{9mm}}}{\textbf{Cell size 15cm}}                                  & \multicolumn{3}{c}{\textbf{Cell size 20cm}}                                  \\[1mm]
                  &               & \textbf{mIoU}$\,{}^\uparrow$ & \textbf{invR}$\,{}^\downarrow$ & \textbf{mPAcc}$\,{}^\uparrow$ & \textbf{mIoU}$\,{}^\uparrow$ & \textbf{invR}$\,{}^\downarrow$ & \textbf{mPAcc}$\,{}^\uparrow$ & \textbf{mIoU}$\,{}^\uparrow$ & \textbf{invR}$\,{}^\downarrow$ & \textbf{mPAcc}$\,{}^\uparrow$ & \textbf{mIoU}$\,{}^\uparrow$ & \textbf{invR}$\,{}^\downarrow$ & \textbf{mPAcc}$\,{}^\uparrow$ \\
\midrule%
\multirow{10}{*}{\textbf{\rotatebox{90}{GT segmentation$\,\;$}}} & S-BKI (10cm)  & 69.18               & 0.00                  & 84.35                 & 65.10               & 0.37                  & 80.60                 & 59.56               & 2.49                  & 74.19                 & 49.16               & 16.63                 & 60.71                 \\
                  & S-BKI (20cm)  & 68.66               & 0.36                  & 83.12                 & 63.53               & 0.36                  & 79.80                 & 60.42               & 0.38                  & 77.07                 & 57.10               & 0.13                  & 72.47                 \\
                  & S-BKI (30cm)  & 68.10               & 0.43                  & 80.05                 & 63.53               & 0.44                  & 77.01                 & 59.75               & 0.45                  & 74.22                 & 55.99               & 0.52                  & 70.65                 \\
                  & S-BKI (40cm)  & 64.15               & 0.57                  & 74.26                 & 59.27               & 0.57                  & 69.53                 & 56.91               & 0.61                  & 67.95                 & 53.27               & 0.71                  & 64.59                 \\
\cmidrule(){2-14}%
                  & OS-BKI (10cm) & 67.91               & 0.33                  & 83.39                 & 64.28               & 0.33                  & 79.94                 & 61.71               & 0.40                  & 76.81                 & 51.88               & 6.75                  & 64.44                 \\
                  & OS-BKI (20cm) & 65.82               & 0.33                  & 81.85                 & 60.95               & 0.33                  & 77.90                 & 59.13               & 0.33                  & 76.58                 & 56.19               & 0.34                  & 72.18                 \\
                  & OS-BKI (30cm) & 66.02               & 0.33                  & 81.52                 & 62.47               & 0.33                  & 79.23                 & 59.27               & 0.33                  & 76.89                 & 55.12               & 0.33                  & 72.27                 \\
                  & OS-BKI (40cm) & 65.82               & 0.33                  & 80.16                 & 63.55               & 0.33                  & 79.08                 & 60.47               & 0.33                  & 76.81                 & 56.13               & 0.33                  & 72.65                 \\
\cmidrule(){2-14}%
                  & \textbf{S-NDT}         & \textbf{78.28}               & 2.93                  & \textbf{88.30}                 & \textbf{72.04}               & 0.83                  & \textbf{85.12}                 & \textbf{66.53}               & 0.73                  & \textbf{81.13}                 & \textbf{61.96}               & 0.78                  & \textbf{78.10}                 \\
\addlinespace[.4em]
\midrule%
\multirow{10}{*}{\textbf{\rotatebox{90}{Predicted segmentation$\;$}}} & S-BKI (10cm)  & 37.95               & 3.23                  & 50.79                 & 35.48               & 3.53                  & 47.77                 & 32.99               & 6.66                  & 44.87                 & 26.89               & 21.88                 & 35.80                 \\
                  & S-BKI (20cm)  & 35.92               & 4.16                  & 47.37                 & 34.14               & 4.56                  & 45.67                 & 33.11               & 4.70                  & 44.62                 & 31.19               & 4.94                  & 42.28                 \\
                  & S-BKI (30cm)  & 33.94               & 5.18                  & 43.99                 & 32.37               & 5.92                  & 42.42                 & 31.37               & 6.10                  & 41.62                 & 29.88               & 5.98                  & 39.59                 \\
                  & S-BKI (40cm)  & 31.55               & 6.08                  & 40.60                 & 28.98               & 7.60                  & 37.22                 & 28.42               & 8.09                  & 36.76                 & 27.11               & 8.26                  & 35.44                 \\
\cmidrule(){2-14}%
                  & OS-BKI (10cm) & 36.78               & 0.33                  & 52.24                 & 35.30               & 0.33                  & 50.61                 & 34.00               & 0.40                  & 49.08                 & 28.58               & 0.66                  & 41.46                 \\
                  & OS-BKI (20cm) & 36.26               & 0.33                  & 51.47                 & 34.27               & 0.33                  & 49.52                 & 33.21               & 0.33                  & 48.86                 & 31.85               & 0.34                  & 46.80                 \\
                  & OS-BKI (30cm) & 36.52               & 0.33                  & 51.26                 & 35.03               & 0.33                  & 50.23                 & 33.55               & 0.33                  & 49.03                 & 31.58               & 0.33                  & 46.35                 \\
                  & OS-BKI (40cm) & 36.53               & 0.32                  & 50.58                 & 35.70               & 0.33                  & 50.40                 & 34.13               & 0.33                  & 49.04                 & 32.51               & 0.33                  & 46.58                 \\
\cmidrule(){2-14}%
                  & \textbf{S-NDT}         & \textbf{41.12}               & 2.93                  & \textbf{56.18}                 & \textbf{39.62}               & 0.83                  & \textbf{54.73}                 & \textbf{37.68}               & 0.73                  & \textbf{52.82}                 & \textbf{35.56}               & 0.78                  & \textbf{50.37}        \\                
\bottomrule[1pt]
\end{tabular}
}
\label{tab:experiments:results}
\vspace{-5mm}
\end{table*}

We first evaluate the mapping approaches in a setting assuming perfect semantic segmentation, i.e., using the provided ground-truth annotation. 
The goal is to determine how well semantics can be represented by the resulting maps.
The results are listed in Tab.~\ref{tab:experiments:results} (top).
Tab.~\ref{tab:experiments:timings} further reports the mean map update rates for each configuration.

With perfect semantic segmentation, S-BKI and the proposed OS-BKI perform similarly.
However, the invalid ratios~(invR) and the mean pixel accuracy~(mPAcc) in Tab.~\ref{tab:experiments:results} show that the reported issue of S-BKI increases as the kernel length increases.
For OS-BKI, the invR remains stable enabling better results for these settings without notably lowering the update rate.
However, both approaches cannot compete with the proposed S-NDT that performs better and is much faster across all cell sizes.
Due to the sub-voxel precise semantics, S-NDT allows the same or even higher semantic map quality while using the next coarser cell size.
Fig.~\ref{fig:experiments:ious} (left) visualizes the IoUs for S-BKI and OS-BKI with a cell size of 10\si{\centi\meter} and a reasonable kernel length of 30\si{\centi\meter} and compares them to S-NDT counterparts with same update rate or same map quality.
The IoUs shows that this statement holds true for almost all classes.

Finally, the results of this set of experiments show that, even with perfect segmentation as input to the mapping stage, the discretization of the environment leads to an unavoidable drop in performance compared to the ground-truth annotation.
Fig.~\ref{fig:experiments:ious} (left) illustrates that a large part of this drop is mainly caused by small or thin objects~(otherprop, lamp, books, shelves, towel, box, sink, paper).
Mapping such classes accurately would require even much finer cell sizes.

\begin{table}[!b]
\vspace{-5mm}
\caption{%
    Mean map update rate~(\si{\hertz}) over all trajectories of the Hypersim test set for all approaches listed in Tab.~\ref{tab:experiments:results}. %
    As the update rate depends on the size of the scene, further, the mean for the slowest (top) and fastest (bottom) scene is reported to the right. %
    The update rate is determined for single-thread processing on our robot featuring an Intel i7 6700HQ mobile CPU. %
    The best result is printed in bold.
}
\vspace{-2.5mm}
\centering
\begin{tabular}{@{}lcccc@{}}
\toprule[1pt]%
              & \multicolumn{4}{c}{\textbf{Mean map update rate}$\,{}^\uparrow$} \\
Cell size     & 5cm                                             & 10cm                                            & 15cm                                            & 20cm                                            \\
\midrule%
\updateRate{S-BKI (10cm)}{}{}  & \updateRate{0.2}{0.1}{3.8} & \updateRate{0.9}{0.3}{8.0} & \updateRate{2.0}{0.7}{10.3} & \updateRate{3.4}{1.2}{11.0} \\
\updateRate{S-BKI (20cm)}{}{}  & \updateRate{0.2}{0.1}{3.6} & \updateRate{0.9}{0.3}{7.9} & \updateRate{2.0}{0.7}{10.0} & \updateRate{3.4}{1.2}{11.1} \\
\updateRate{S-BKI (30cm)}{}{}  & \updateRate{0.2}{0.1}{3.6} & \updateRate{0.8}{0.3}{7.9} & \updateRate{2.0}{0.6}{10.1} & \updateRate{3.4}{1.2}{11.0} \\
\updateRate{S-BKI (40cm)}{}{}  & \updateRate{0.2}{0.1}{3.7} & \updateRate{0.8}{0.3}{7.8} & \updateRate{1.9}{0.6}{9.3} & \updateRate{3.3}{1.2}{11.1} \\
\midrule%
\updateRate{OS-BKI (10cm)}{}{} & \updateRate{0.2}{0.1}{3.8} & \updateRate{0.9}{0.3}{7.6} & \updateRate{1.9}{0.6}{9.0} & \updateRate{3.5}{1.3}{11.2} \\
\updateRate{OS-BKI (20cm)}{}{} & \updateRate{0.2}{0.1}{3.8} & \updateRate{0.9}{0.3}{8.1} & \updateRate{1.9}{0.6}{9.3} & \updateRate{3.1}{1.1}{10.6} \\
\updateRate{OS-BKI (30cm)}{}{} & \updateRate{0.2}{0.1}{3.6} & \updateRate{0.9}{0.2}{7.9} & \updateRate{1.9}{0.6}{9.5} & \updateRate{3.0}{1.1}{9.8} \\
\updateRate{OS-BKI (40cm)}{}{} & \updateRate{0.2}{0.1}{3.8} & \updateRate{0.8}{0.2}{7.2} & \updateRate{1.9}{0.6}{9.7} & \updateRate{3.0}{1.1}{9.9} \\
\midrule%
\updateRate{\textbf{S-NDT}}{}{}         & \updateRate{\textbf{1.0}}{0.8}{1.2}  & \updateRate{\textbf{3.5}}{3.0}{6.8}  & \updateRate{\textbf{5.2}}{4.6}{14.9} & \updateRate{\textbf{5.4}}{4.8}{21.9} \\
\bottomrule[1pt]
\end{tabular}
\label{tab:experiments:timings}
\end{table}

\subsection{Mapping with Predicted Segmentation}
\label{sec:experiments:segmentation}
Next, we use the output of our ESANet-R34-NBt1D as input for mapping.
This allows comparing the approaches in a more realistic scenario.
The ESANet-R34-NBt1D~(without mapping) reaches a mIoU of 41.17 on the test set.

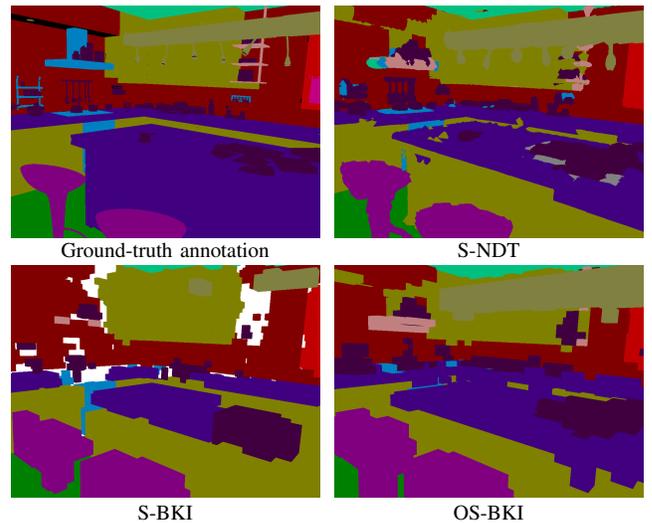
\begin{figure}[!b]
    \vspace{-7mm}
	\centering
	\begin{tikzpicture}[scale=1.0]
	    \node at (0, 0){%
	        \includegraphics[width=4.1cm,trim=0cm 0cm 0cm 0cm, clip]{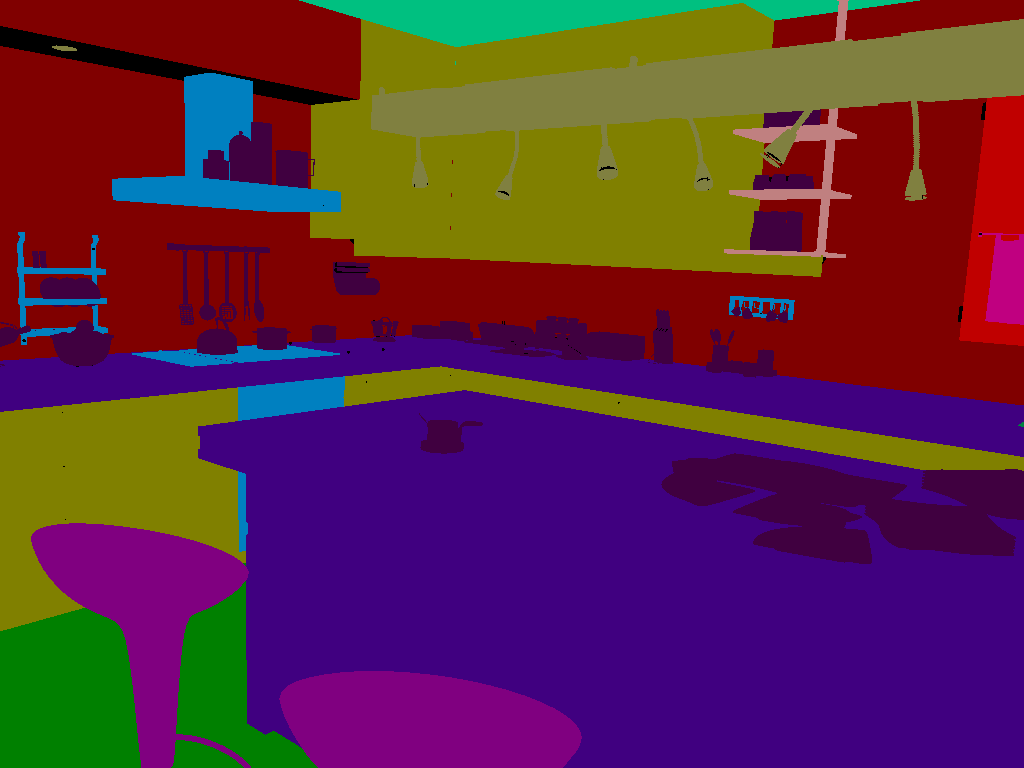}%
	    };
	    \node at (4.3, 0){%
	        \includegraphics[width=4.1cm, trim=0cm 0cm 0cm 0cm, clip]{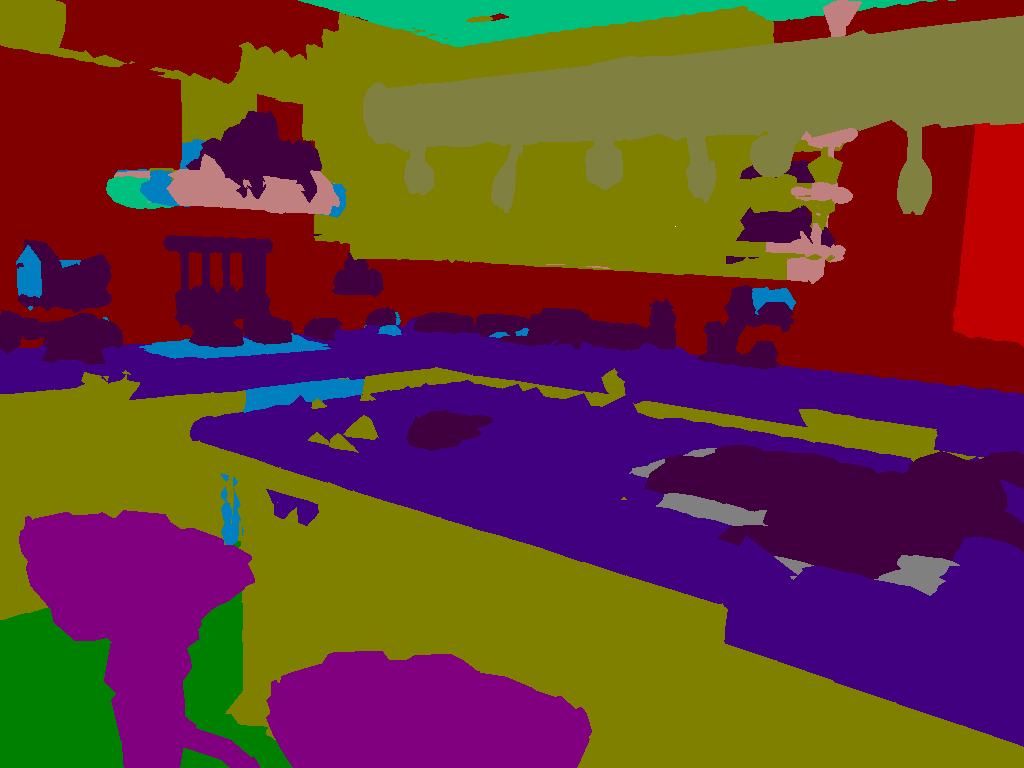}%
	    };
	    \node at (0, -3.45){%
	        \includegraphics[width=4.1cm, trim=0cm 0cm 0cm 0cm, clip]{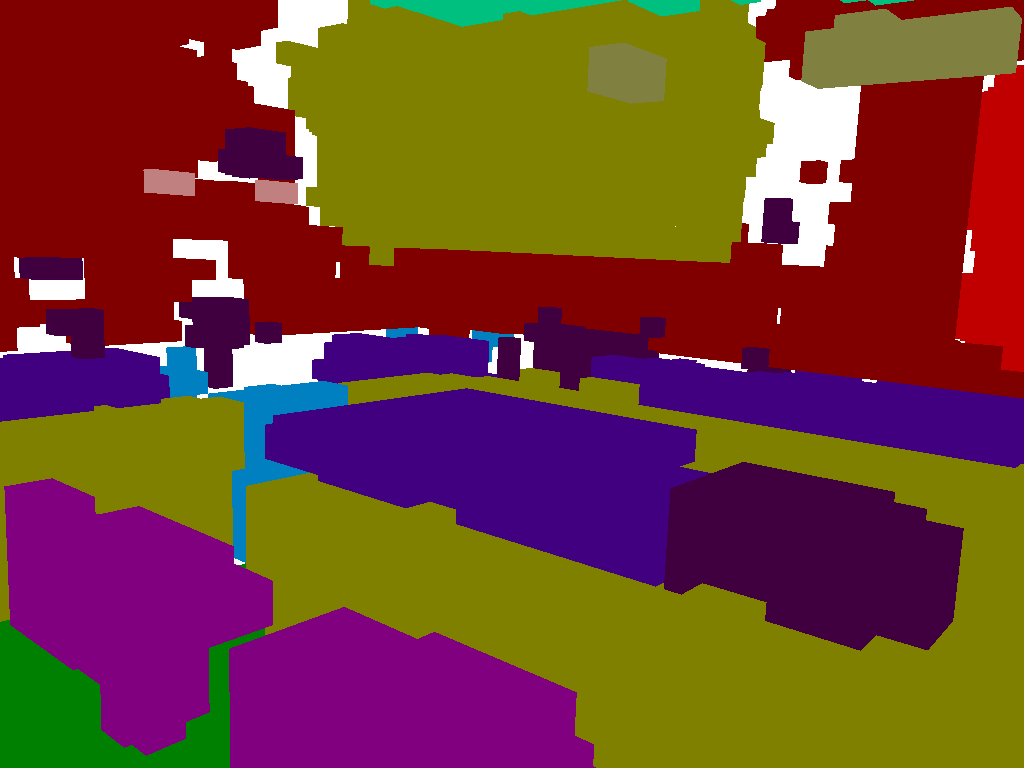}%
	    };
	    \node at (4.3, -3.45){%
	        \includegraphics[width=4.1cm, trim=0cm 0cm 0cm 0cm, clip]{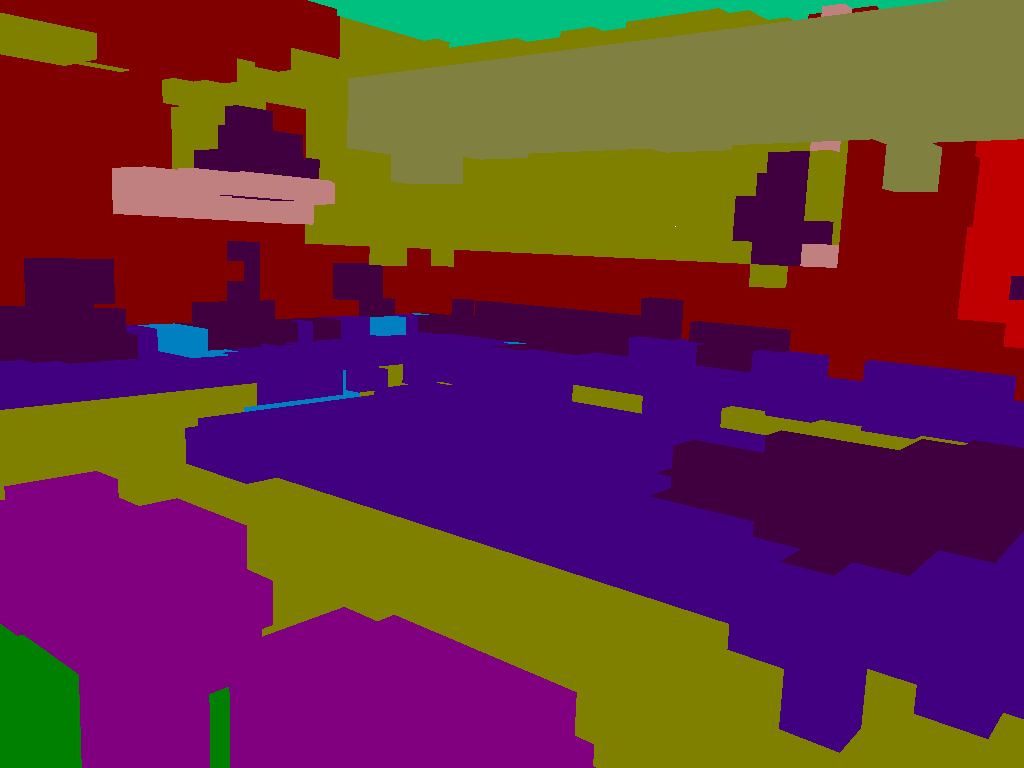}%
	    };
	    \node[anchor=north] at (0,-1.48) {\footnotesize Ground-truth annotation};
	    \node[anchor=north] at (4.3,-1.48) {\footnotesize S-NDT};	    
	    \node[anchor=north] at (0,-4.98) {\footnotesize S-BKI};
	    \node[anchor=north] at (4.3,-4.98) {\footnotesize OS-BKI};
	\end{tikzpicture}%
	\vspace{-4mm}
    \caption{
        Ground-truth annotation and back-projections for the same camera pose of scene \emph{ai\_001\_010/cam\_01} of the Hypersim test set when mapping with predicted segmentation and cell size of 10\si{\centi\meter}. %
        For S-BKI and OS-BKI, a kernel length of 30\si{\centi\meter} is used. %
        White indicates invalid pixels (free space), black pixels belong to the void class. %
        See Fig.~\ref{fig:experiments:ious} for remaining colors.
    }%
    \label{fig:experiments:backprojection}
\end{figure}

As shown in Tab.~\ref{tab:experiments:results}~(bottom) and further visualized by the back-projections in Fig.~\ref{fig:experiments:backprojection}, unlike our proposed OS-BKI, S-BKI does not allow satisfying mapping in this scenario. 
In areas where multiple labels occur, free space tends to be over-presented resulting in holes in the map (invalid pixels in the back-projection).
For OS-BKI, the ratio of invalid pixels is smaller, leading to much better results.

\begin{figure*}[!t]
    \vspace{1mm}%
    \hspace{-2mm}
    \begin{tikzpicture}[scale=1.0]%
	    \node[anchor=north west] at (-0.1, 0){%
	        \includegraphics[trim=0.2cm 0.3cm 0.25cm 0.25cm, clip, width=0.47\textwidth]{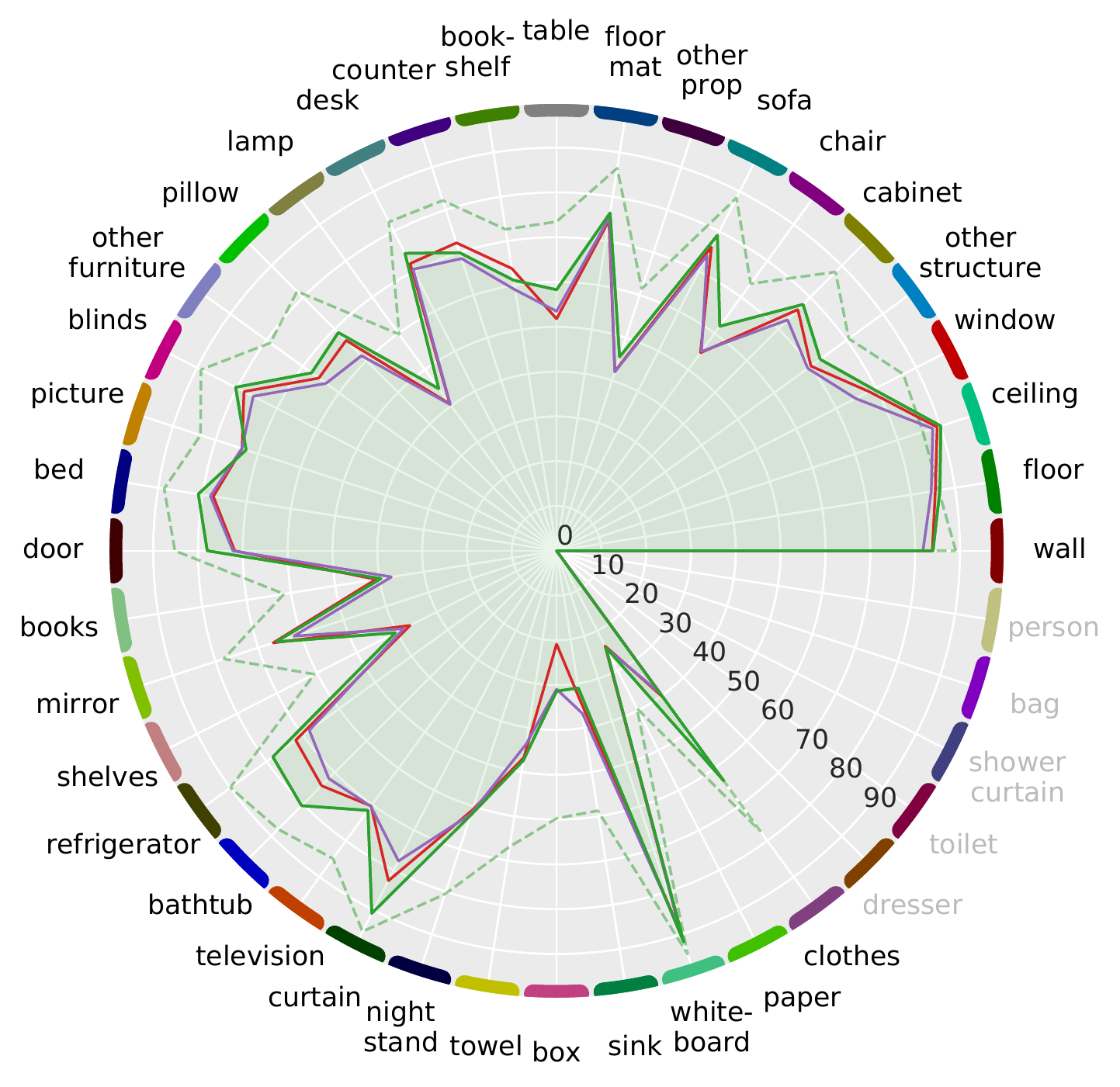}%
	    };
	    \node[anchor=north west] at (9.3, 0){%
	        \includegraphics[trim=0.2cm 0.3cm 0.25cm 0.25cm, clip, width=0.47\textwidth]{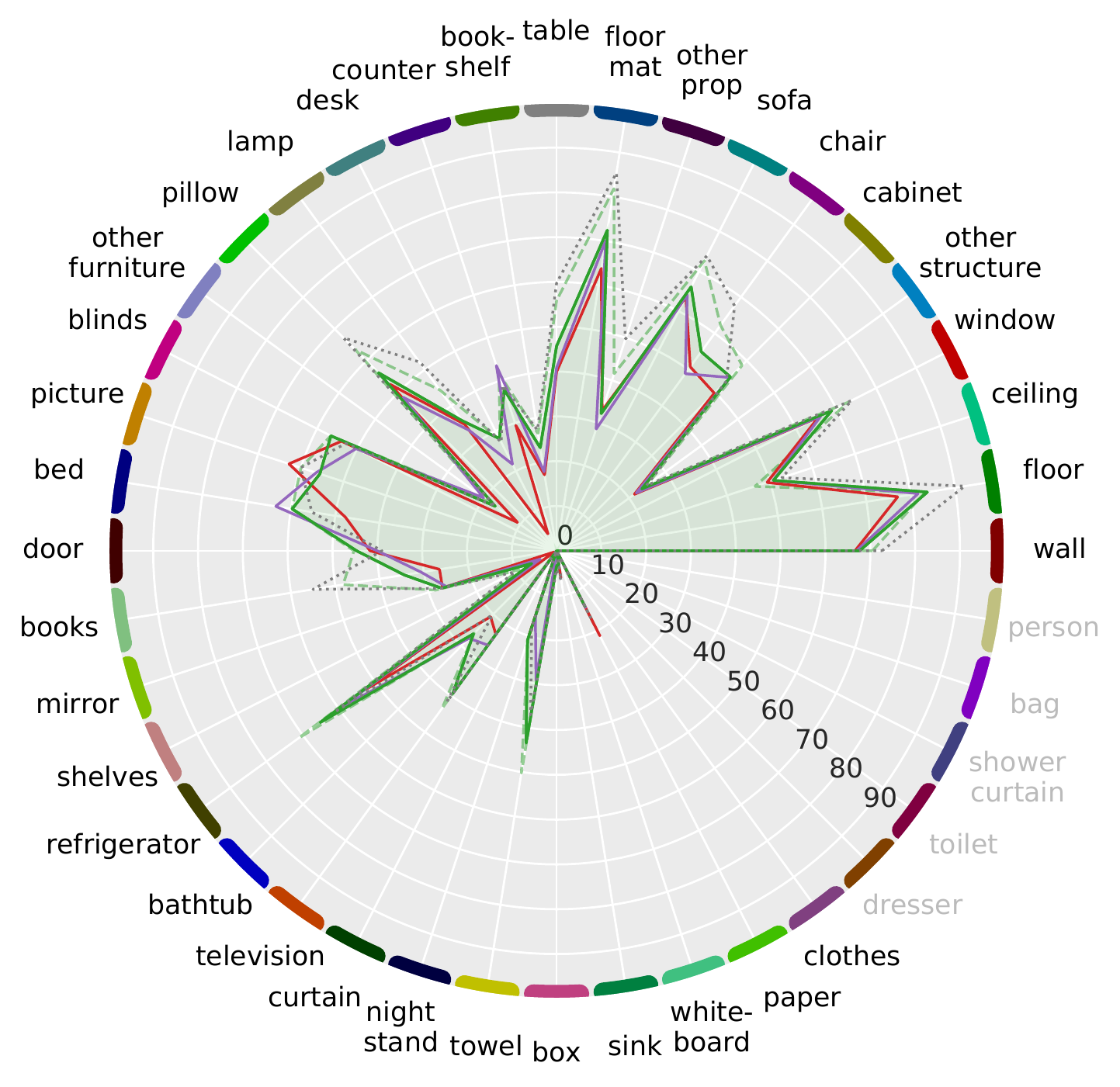}%
	    };%
	    \node[anchor=north west, align=left] at (0, 0) {\footnotesize\bf Ground-truth\\\footnotesize\bf segmentation};%
	    \node[anchor=north west, align=left] at (8.9, 0) {\footnotesize\bf Predicted\\\footnotesize\bf segmentation};%
        \node[anchor=center] at (9.1, -6.9) {%
	        \includegraphics[trim=4.7cm 14.1cm 4.25cm 0.25cm, clip, width=0.135\textwidth]{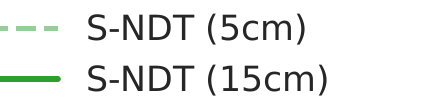}%
	    };%
        \node[anchor=center] at (8.85, -7.55) {%
	        \includegraphics[trim=4.7cm 12.9cm 4cm 1.3cm, clip, width=0.145\textwidth]{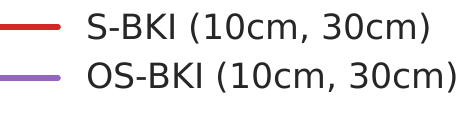}%
	    };%
        \node[anchor=center] at (8.85, -7.95) {%
	        \includegraphics[trim=4.7cm 12.5cm 3.5cm 2.3cm, clip, width=0.165\textwidth]{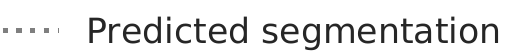}%
        };%
    \end{tikzpicture}
    \vspace{-9mm}
    \caption{%
        Per-class intersection over union (IoU) on the Hypersim test set when mapping with the ground-truth segmentation~(left) and predicted segmentation of ESANet~(right). %
        Classes are ordered descending by frequency counterclockwise. %
        Classes printed in gray do not appear in the test set. %
        Cell size and kernel length (2nd parameter) are denoted in brackets.%
    }
    \label{fig:experiments:ious}
    \vspace{-4mm}
\end{figure*}

However, again, S-NDT performs much better, almost reaching the performance of the ESANet-R34-NBt1D for a cell size of 5\si{\centi\meter}.
The IoUs in Fig.~\ref{fig:experiments:ious}~(right) further indicate that the results are close for almost all classes.
Moreover, for some classes~(cabinet, counter, blinds, bed, door, refrigerator, bathtub,~\ldots) integrating predictions over time in semantic NDT maps seems to improve results.

The results in Tab.~\ref{tab:experiments:results} for both experiments on Hypersim show that the semantic quality of maps created with S-NDT is mainly limited by the quality of the preceding semantic segmentation.

\section{Application in Real-World Scenario}
\label{sec:application}

\begin{figure}[!b]
    \vspace{-8mm}
    \centering
    \begin{tikzpicture}
	    \node at (0, 0){%
	        \includegraphics[width=0.98\columnwidth, trim=0.6cm 1cm 0 2cm, clip]{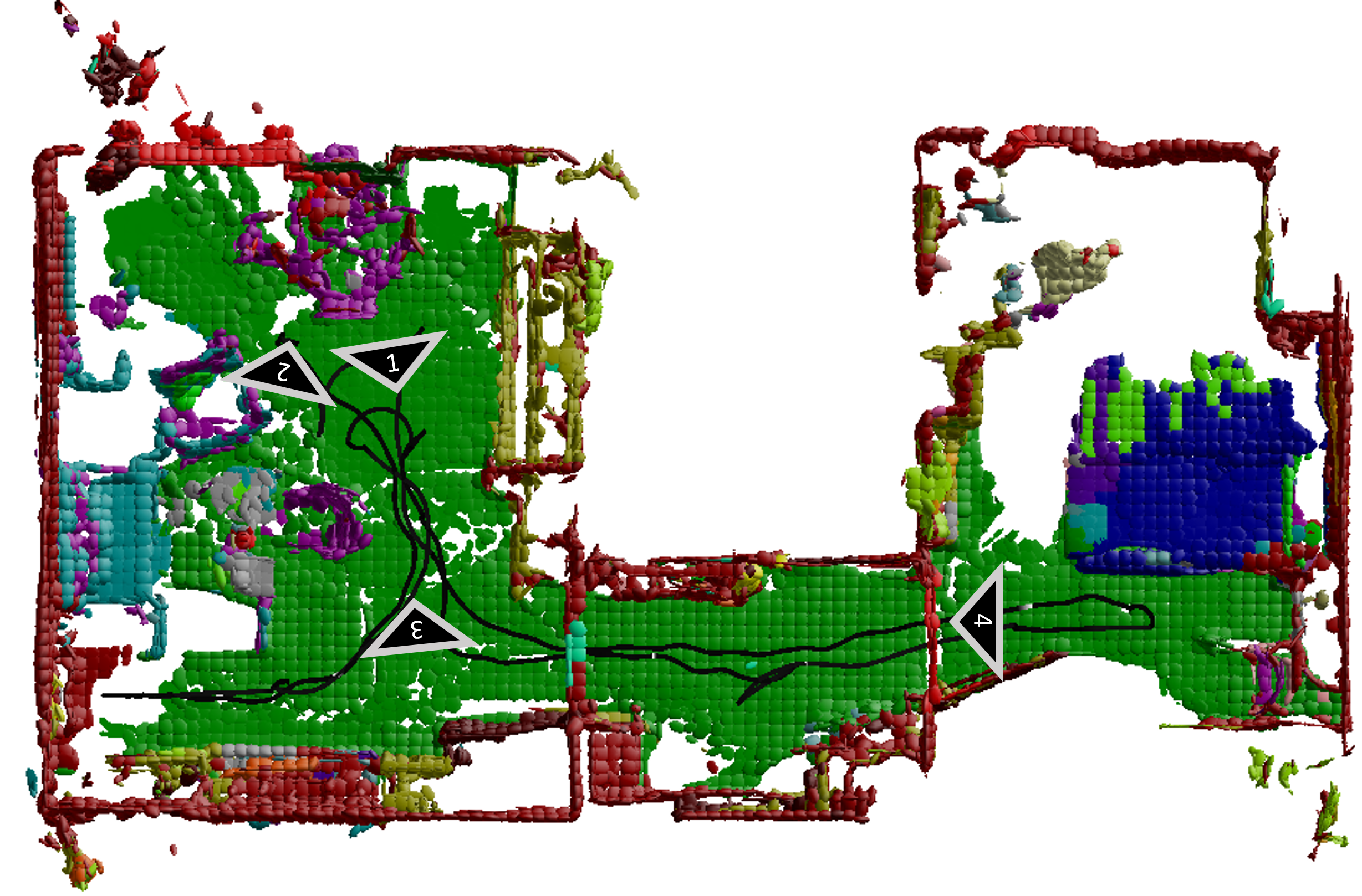}%
	    };
	    \node at (0, -5.5){%
	        \includegraphics[width=0.97\columnwidth]{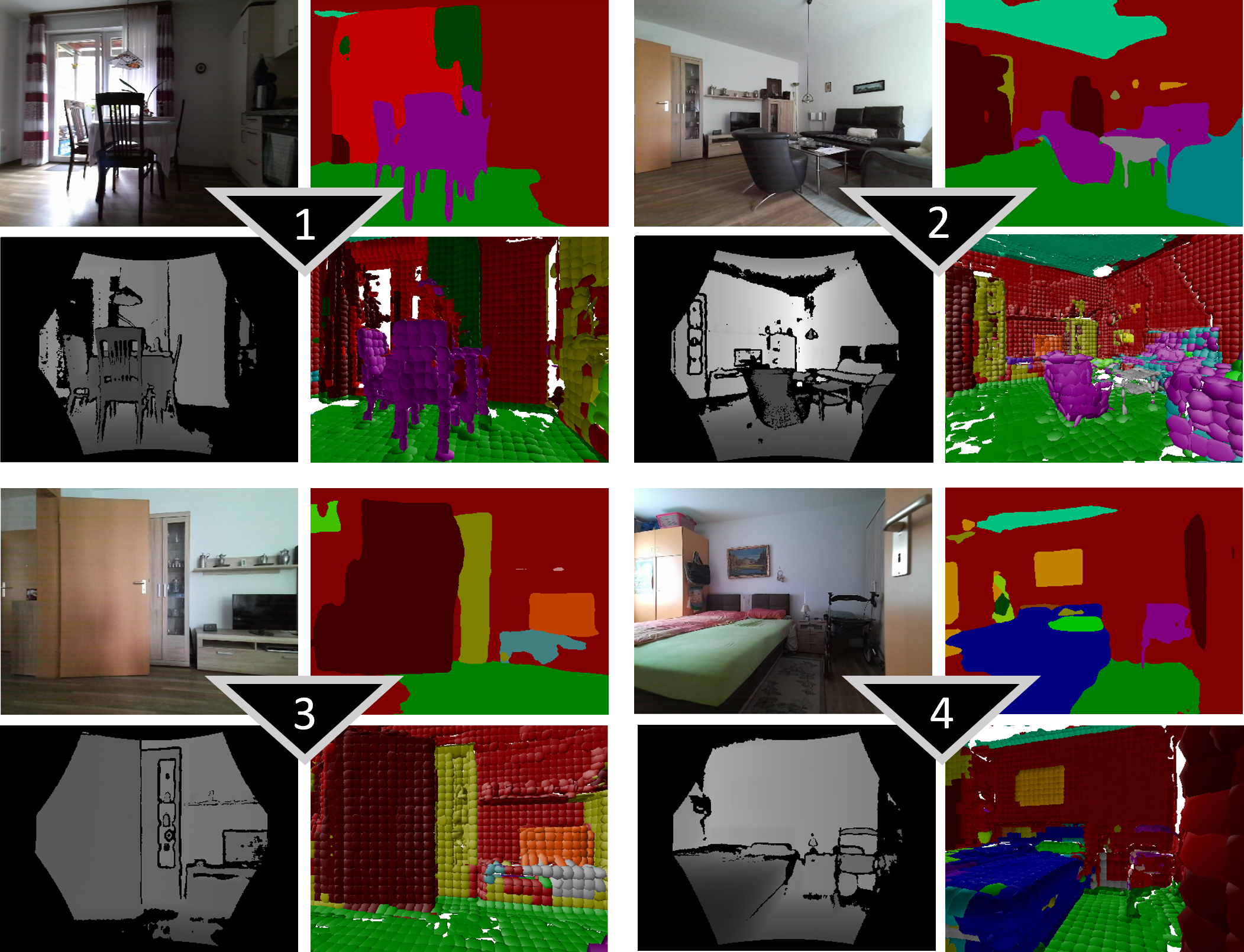}%
	    };
    \end{tikzpicture}%
    \vspace{-4mm}
    \caption{%
        Top: Bird's-eye view of the S-NDT map of the apartment of a test participant including the robot's trajectory (black). %
        Bottom: for selected poses (1-4), clockwise from bottom left: depth and color image, semantic segmentation of ESANet-R34-NBt1D, and 2D rendering visualizing the map from the robot's pose. %
        Best viewed at 200\%, see Fig.~\ref{fig:experiments:ious} for label colors.
    }%
    \label{fig:application}
    \vspace{-1mm}
\end{figure}

We apply our proposed two-step pipeline for efficient and robust semantic mapping in the context of our MORPHIA project in various apartments of seniors, such as the one shown in Fig.~\ref{fig:application}.
After an initial non-semantic mapping phase using RTAB-Map~\cite{Labbe-JFR-2019}, it is switched to localization mode to provide a reliable long-term localization.
For short-term semantic mapping, we choose the ESANet-R34-NBt1D trained on SUNRGB-D~\cite{Song-CVPR-2015-SUNRGBD} and S-NDT with cell size of 10\si{\centi\meter}.
Even though SUNRGB-D does not contain any images of our Azure Kinect sensor setup, the mapping results show a suitable level of quality, modeling the semantics of the environment as well as dynamic objects, such as chairs, doors, or clothes, along with their spatial extend.
This enables our robot to be aware of the current scene and to perform high-level semantic control commands or delivery tasks for the senior.
For further impressions on the semantic mapping, we refer to the attached video or our repository on GitHub.

\section{Conclusion and Future Work}
\label{sec:conclusion}
In this paper, we have incorporated semantic information into efficient occupancy NDT maps realizing semantic occupancy NDT (S-NDT) maps. 
Our experiments on the Hypersim dataset~\cite{Roberts-ICCV-2021-Hypersim} revealed that S-NDT mapping outperforms the recent state-of-the-art approach S-BKI and an optimized version of it in terms of both performance and runtime.
Our code for preparing the Hypersim dataset, the weights for our applied ESANet-R34-NBt1D, and the code for evaluation are publicly available.
Thereby, we hope that we can call the community's attention to the Hypersim dataset for evaluating semantic mapping approaches.

We also showed that S-NDT mapping combined with our ESANet-R34-NBt1D represents a powerful instrument for real-world robotic applications, enabling mobile robots to gain a strong understanding of their environment.

In future work, we intend to conduct experiments on further datasets such as ScanNet~\cite{dai2017scannet} as well.%

\ifthenelse{\boolean{isarxiv}}{%
}{%
    \addtolength{\textheight}{-5cm}   %
}%

\bibliographystyle{bib/IEEEtran}
\bibliography{bib/literature}

\end{document}